# Agreement Maintenance Based on Schema and Ontology Change in P2P Environment

Lintang Y. Banowosari, I Wayan S. Wicaksana, A. Benny Mutiara
*Gunadarma University*
{lintang,iwayan,amutiara}@staff.gunadarma.ac.id

**Abstract**

*This paper is concern about developing a semantic agreement maintenance method based on semantic distance by calculating the change of local schema or ontology. This approach is important in dynamic and autonomous environment, in which the current approach assumed that agreement or mapping in static environment.*

*The contribution of this research is to develop a framework based on semantic agreement maintenance approach for P2P environment. This framework based on two level hybrid P2P model architecture, which consist of two peer type: (1) super peer that use to register and manage the other peers, and (2) simple peer, as a simple peer, it exports and shares its contents with others. This research develop a model to maintain the semantic agreement in P2P environment, so the current approach which does not have the mechanism to know the change, since it assumed that ontology and local schema are in the static condition, and it is different in dynamic condition. The main issues are how to calculate the change of local schema or common ontology and the calculation result is used to determine which algorithm in maintaining the agreement.*

*The experiment on the job matching domain in Indonesia have been done to show how far the performance of the approach. From the experiment, the main result are (i) the more change so the F-measure value tend to be decreased, (ii) there is no significant different in F-measure value for various modification type (add, delete, rename), and (iii) the correct choice of algorithm would improve the F-measure value.*

## 1. Introduction

Internet has contributed great value for data exchange. On other hand, Internet introduced some new issues. Currently, information sources are more massive, distributed, dynamic and open. Many people have become accustomed to the Internet's rapid growth. One function of Internet is for searching and sharing information. All existed information of Internet kept by various data sources. Many sources sometimes present information at different model databases, including highest level until lower level. Every level needs different kind, attribute and properties which can be saved in database. The differences data sources can make problem in accessing information in different sources, especially when implemented in network model, for example P2P (Peer to Peer).

Recently, the computer science community has become accustomed to the Internet's continuing rapid growth, but even to such jaded observers the explosive increase in Peer-to-Peer (P2P) network usage has been astounding [6].

In this paper we focus to solve problem for accessing information in different P2P sources especially in bridging query from user to peer with similar property or object. To solve that problem we use method of semantic agreement maintenance in implemented with semantic web. The goal of this paper is to reduce problem in accessing information with our offered method and implemented method to the web services.

Main motivations of our approach as follow: we divide it for user and system. For user, source of information contains various models to represent their content. Problem occurs when they want to get information from different databases. We propose a new approach to solve that problem, so for retrieval user can get more relevant information. For system, we hope we can bridges the differences between databases by semantic agreement maintenance. So we can minimize manual monitoring, which high failure and high cost. And finally we can deliver an automatic monitoring and improvement agreement idea.

The creation of semantic mappings between different information sources is the crucial point in integration approach. Many existing studies are based on the idea that the mappings can easily be created by expert designers when the schema of the different information sources are combined and integrated. To create mappings between semantically related concepts of heterogeneous sources, a number of different criteria can be used among which the most obvious is matching the names of schema elements. Linguistic comparison methods can be used to match the names or labels of schema element by relying on their latent semantic. On a higher level, the structure of the information can be used as a criterion (e.g. the attributes of a class) for identifying related concepts.

In the dynamic environment, local schema and ontology can be changed or updated. When there is some query request for information of that system, then the query result is not valid, since the agreement that has been formed before is still the same. So that it can not fulfill the concept of the community member. For that reason it can be done the monitoring by network administrator manually to know whether there is some changed on the local schema and ontology, but it is time consuming and it is difficult to predict the number of the peer that is changed and the number its changed.

Although it is often happen that there is some invalid mapping can cause the integration system failed, but in fact that there is a little research in mapping maintenance. Currently, the integrated system mostly still maintain the mapping manually, in a process which is expensive and possibility error occurred is big. Therefore, the more efficient solution is needed for significantly to reduce the data integration cost.

Problems in developing semantic agreement maintenance method are:

(1) to detect changes of data sources. Some modification can be founded in data source during data operations.

(2). to compare each version of data source modification to count how big changes of it by giving value to each operation (see General Overview for details).

(3). to choose algorithm that we'll used for maintain data source, after we get the total value of its operations (border value).

The purpose of paper is to present a semantic agreement maintenance for solve problem accessing information in different data sources and give information about semantic web which can be solve the problem for accessing information and developing one modification approach of distance semantic theory to know change of local scheme or ontology so that can be used to conduct conservancy of agreement between a common ontology and of provider existing peer.

## 2. Approaches Review

To answer user queries, a data integration system employs a set of semantic mappings between the mediated schema and the schema of data sources. In dynamic environments sources often undergo changes that invalidate the mappings. Hence, once the system is deployed, the administrator must monitor it over time, to detect and repair broken mappings. Today such continuous monitoring is extremely labor intensive, and poses a key bottleneck to the widespread deployment of data integration systems in practice.

One approach is Maveric [5], an automatic solution to detecting broken mappings. At the heart of Maveric is a set of computationally inexpensive modules called sensors, which capture salient characteristics of data sources (e.g., value distributions, HTML layout properties). Maveric trains and deploys the sensors to detect broken mappings. Maveric also have three novel improvements: perturbation (i.e., injecting artificial changes into the sources) and multi-source training to improve detection accuracy, and filtering to further reduce the number of false alarms.

The other approach is working in XML p2p database systems [4]. This approach presented a novel technique for detecting corrupted mappings in XML p2p data integration systems. This technique can be used in any context where a schema mapping approach is used, and it is based on a semantic notion of mapping correctness, unrelated to the query transformation algorithms being used. This form of correctness works on the ability of a mapping to satisfy the target schema, and it is independent from queries.

Semantic integration is an active research in several disciplines, such as databases, information, integration, and ontologies and to represent mapping ontology we can use several tools, one of the tools is PROMPT, and the tools are extensions to the Protege ontology-development environment [4]. Semantic similarity relates to computing the between concept which are not lexicographically similar. Some of the most popular semantic similarity methods are implemented and evaluated using WordNet as the underlying reference ontology [2].

## 3. Proposed Methodology

Semantics is the study of language meaning. In the computer science, semantics have meaning of program or function. Semantics is growing up and become Semantic Web which development of World Wide Web through implant with semantic metadata [2]. Semantic conflict arise when two system do not use same interpretation of information. Simplest form of disagreement in interpreting information is homonym (using word which is equal to different meaning), and synonym (using word differ from is same meaning). In this case, semantics of information
Have to be considered by for the agenda of deciding how different information item correlates one with the other. Yaser [1] have divided schematic variety to in a few groups: Differing in class like synonym, homonym, differing in class attribute, integrity constrain and method.

Differing in attribute, like domain, unit, assess data type and default Differ in hierarchy, like class, attribute, generalizing storey; level and of aggregation.

One approach with semantic agreement maintenance is using four steps to detect and determine whether it needs maintenance or not (see figure 1). The main steps as follow:

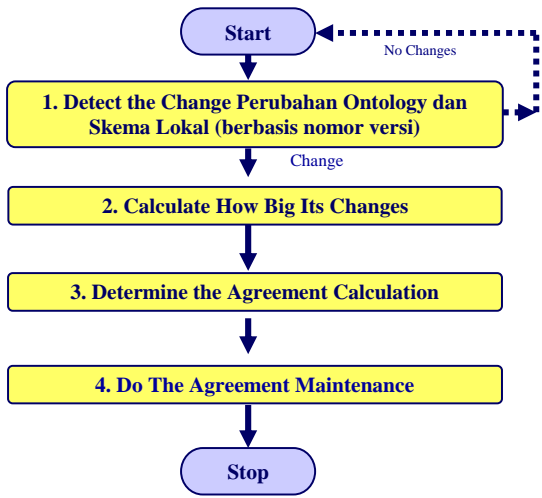

Figure 1 General Flow in Agreement Maintenance

1. Detecting the changes of Common Ontology and Local Scheme; for detail step, see figure 2. The changes of one Common Ontology and Local Scheme can be detected by Common Ontology versioning [1]. Common Ontology versioning builds and maintains different version from Common Ontology and provides access for them. Currently, versioning mechanism doesn't support log of changes. The common approaches of versioning are implemented in DNS master-slave and CVS software development repository. Fortunately, OWL support information about versioning.

2. Calculating how big its changes, for the detail step see figure 3. Referring to the point 1 above, a mechanism is needed to solve the weakness in unavailable log of changes. We purpose a mechanism to calculate how big the changes between previous and current version. The changes of Common Ontology can utilize PROMPTDIFF algorithm, that introduced by Noy and Klein [3].

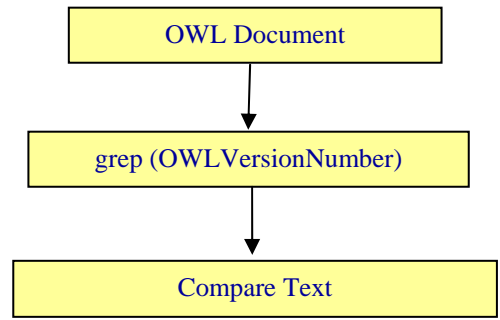

Figure 3 Version Number Detection

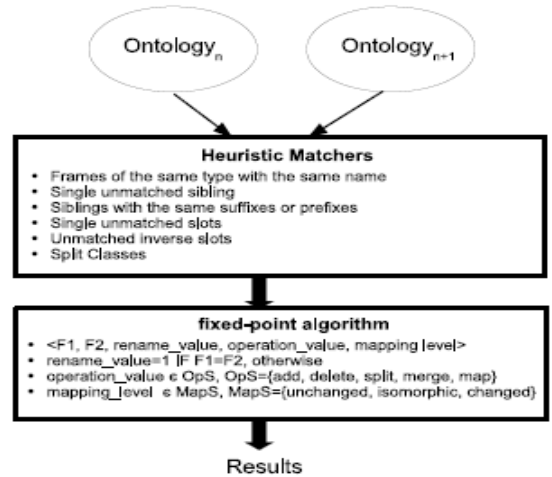

Figure 4 PromptDiff Algorithm

3. Determine the Agreement Calculation Algorithm. Choosing maintenance algorithm based on border value of changing calculation. The border value can be found based on empirical trial-error approach. The border value can be implemented to decide which algorithm will be conducted. See the figure 5.

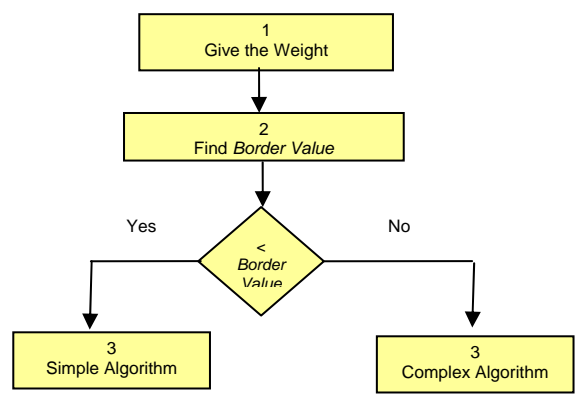

Figure 5 PromptDiff Algorithm

4. Do the agreement maintenance. We consider two types of algorithm for doing the agreement maintenance: simple and complex. The complex algorithm is algorithm which includes label matching with Jiang & Conrath, internal matching, and external matching, see the figure 6 (b). The simple algorithm is a part of complex algorithm, which only has one step in label matching using Jiang & Conrath see the figure 6 (a).

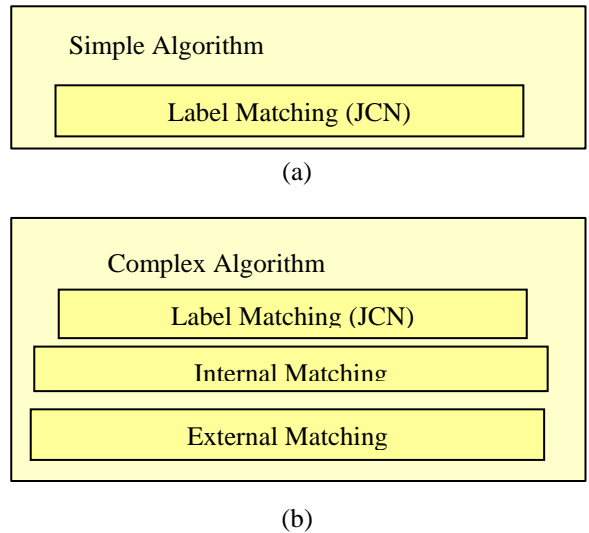

(a)

(b)

Figure 6 Agreement Maintenance Algorithm

The JCN equations are as follows:

$$Sim_{label} = sim_{jcn}(c_1,c_2) = \max\left[\frac{1}{dist_{jcn}(c_1,c_2)}\right]$$

Where $dist_{jcn}(c_1,c_2) = IC_{c1} + IC_{c2} - 2 * IC(LCS(c_1,c_2))$, LCS is the lowest node that subsumes or dominates $c1$, $c2$. For instance, *animal* is the lowest common node of *cat* and *dog*. The information content values in equation above are calculated by

$$IC_c = -\log p(c); if p(c) > 0$$

where $c$ is a concept in WordNet and $p(c)$ is the probability of encountering $c$ in a given corpus. The $p(c)$ is defined by:

$$p(c) = \frac{\sum_{w \in W(c)} count(w)}{N}$$

where $W(c)$ is set of words (nouns) in the corpus whose sense are subsumed by concept $c$, and $N$ is the total number of word (noun) tokens in the corpus that are also present in WordNet.

The structures of the compared concepts are used to fine tune the similarity measurement. Two types of structure are considered. The internal structure is compared to the attributes of the concept while the external structure takes into account the relation of a concept to the other concepts of the hierarchy.

## 4. Result and Discussion

There is some preparation before doing the testing, i.e. create the modification of 10 local schemas suitable with the scenario and HR-XML as common ontology [7]. Creation of local schema refer to the domain of testing such that job matching service which has two main component, job seeker and job provider and it comes from 10 peers. In this testing, local schema is created by using Protégé software which has capability to present its schema by using OWL language. And for common ontology, we take from available ontology for HRD domain i.e. HR-XML and that ontology rewrite using Protégé. In executing the testing, it needs some support software tools such as Protégé, PromptTab, and online application for doing the semantic similarity calculation based on WordNet.

We implemented some scenario to test the semantic agreement model by doing the modification to one or more scheme. First of all, we modify (add, delete, and change) the local scheme of peer for its class and property. The modification of local scheme gives a border value. By using a border value, we can take the PromptDiff [3] approach (from PROMPT TAB tool) to find out algorithm that we will use. To test the algorithm that we use, we can make an agreement from two algorithms. And finally, we count the Recall Value (The proportion of relevant document, beyond all exist relevant document), Precision Value (The proportion of retrieved and relevant document for all retrieved document), and F-Measure (harmonic average weight from precision and recall value.

From those scenarios, we got the result of agreement maintenance as shown in figure 7 until 12.

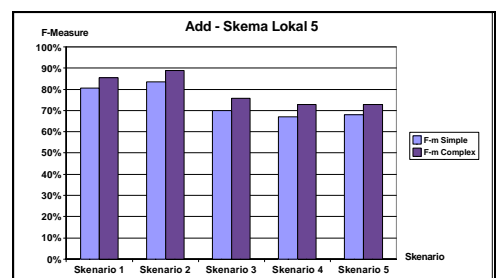

Figure 7 Result on Add Modification LS 6

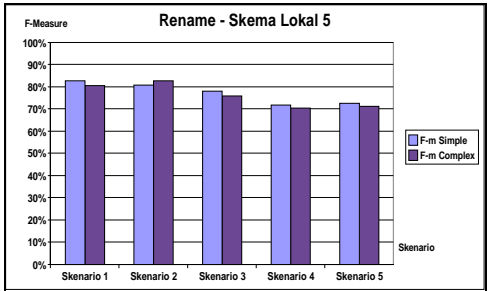

Figure 8 Result on Rename Modification LS 6

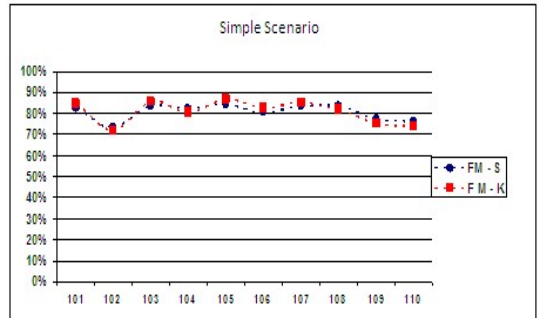

Figure 12 the result of simple algorithm

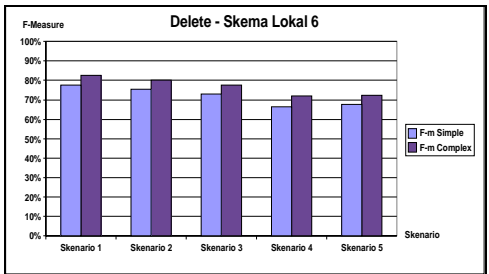

Figure 9 Result on Delete Modification LS 6

The discussions of the results are:

1. Figure 7 until 9 (part of 30 graph figures) are graphs which depict the F-measure values based on modification differences (scenario n). On the figure 7 is Add modification for the local schema 6 (peer 6), figure 8 is Rename modification for the local schema 6 (peer 6), figure 9 is Delete modification for the local schema 6 (peer 6). From the 3 figures, it is shown that the more modification (both for Add, Delete and Rename) the F-measure has tended to decrease.

2. Figure 10 give the information that agreement maintenance has F-measure result which relatively similar for various local schemas (peers). On the other words, this approach has not depended from information sources or peers.

3. Figure 7 until 10 are an analysis for showing that semantic maintenance approach which face with various modification on addition, deletion and rename. We can say that F-measure result for facing various modifications relatively similar for Add, Delete and Rename.

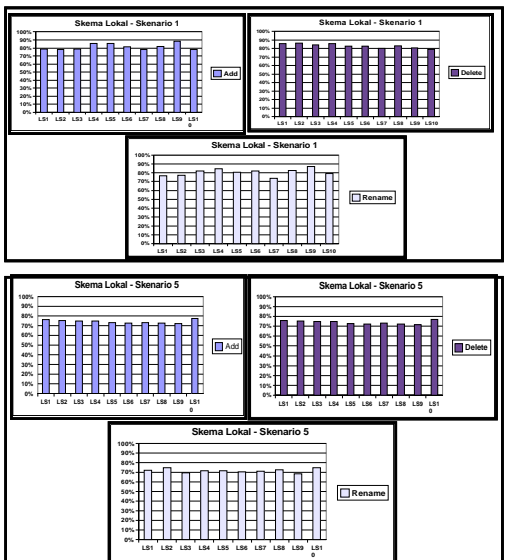

Figure 10 Result experiment on Add, Delete and Rename in LS 1 and 5

4. If the calculation of differences between new and old version of the same Local Scheme and Common Ontology should use the complex algorithm, so the F-measure of complex algorithm is always greater than the F-measure of simple algorithm, see figure 11.

5. Otherwise, if the calculation of differences between new and old version of the same Local Scheme and Common Ontology should use the simple algorithm, the F-measure of complex algorithm is relatively equal to the F-measure of simple algorithm. In many cases, the F-measure of simple algorithm is approximately 2% better than the F-measure of complex algorithm, see figure 12.

From the last two points, so the precise selection on the algorithm is very important to improve the F-measure and also to save the computation cost.

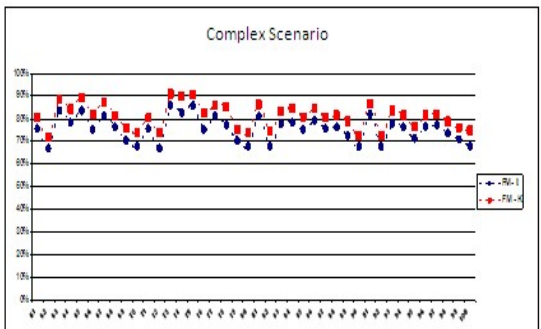

Figure 11 the result of complex algorithm

## 5. Conclusion and Future Work

The more modification of local scheme and ontology makes F-Measure for semantic agreement become worst. Choosing a proper algorithm is the most important thing for agreement maintenance respect to cost computation and F-Measure. The using of complex algorithm for case that should be use a simple algorithm makes higher complexity with relatively small F-measure distinction. Otherwise, the F-measure value of complex algorithm higher than the F-measure value of simple algorithm with higher complexity.

So, simple state that selecting the appropriate algorithm for semantic agreement maintenance is important to get better F-Measure and it is possible to reduce the cost of computing. This statement has been figured out based on the result of evaluation.

Ideally, the semantic agreement maintenance approach is a generic approach. Therefore, the further evaluation for other domains is needed. Furthermore, there are also possibilities to divide value of difference between Local Scheme and Common Ontology to some regions and implement the different algorithm for each region.